\documentclass[twoside]{article}      
\usepackage{amsmath,amssymb,amsfonts,amsthm} 
\usepackage{graphics}                 
\usepackage{color}                    
\usepackage{hyperref}                 
\usepackage{comment} 			
\usepackage{enumitem}

\theoremstyle{plain}
\newtheorem{theorem}{Theorem}[section]
\newtheorem{lemma}[theorem]{Lemma}

\newtheorem{corollary}[theorem]{Corollary}

\theoremstyle{definition}
\newtheorem{definition}{Definition}[section]

\theoremstyle{definition}
\newtheorem{exmp}{Example}[section]

\def\B{\mathbb{ B}}
\def\M{\mathfrak{ M}}
\def\C{\mathfrak{C}}

\def\mm{\mathcal{M}}

\date{\small\it \today}
\title{Sampling and Learning for Boolean Function
\footnote{Great thanks for whole heart support of my wife. Thanks for Internet and research contents contributers to Internet. }}  

\author{ Chuyu Xiong \\
{\small Independent researcher, New York, USA} \\
{\small Email: chuyux99@gmail.com}
}

\begin{document}
\maketitle
\begin{abstract}
In this article, we continue our study on universal learning machine by introducing new tools. We first discuss boolean function and boolean circuit, and we establish one set of tools, namely, fitting extremum and proper sampling set. We proved the fundamental relationship between proper sampling set and complexity of boolean circuit. Armed with this set of tools, we then introduce much more effective learning strategies. We show that with such learning strategies and learning dynamics, universal learning can be achieved, and requires much less data.

\end{abstract}


{\it It can scarcely be denied that the supreme goal of all theory is to make the irreducible basic elements as simple and as few as possible without having to surrender the adequate representation of a single datum of experience.} \hspace{15pt}  \hfill--- A. Einstein \\
\vskip -12pt
{\it...... then a sudden leap takes place in the brain in the process of cognition,......} \hspace{15pt} \hfill--- Mao Zedong \\

\section{Introduction}
In \cite{paper1, paper2, paper4, paper5}, we tried to study universal learning machine. There, we laid out framework of discussions and proved some basic yet important results, such as: with sufficient data, universal learning machine can be achieved. The core of universal learning machine is X-form, which turns out to be a form of boolean function. We showed that the learning is actually equivalent to dynamics of X-form inside a learning machine. Thus, in order to study universal learning machine well, we need to study thoroughly X-form and the motion of X-form under driven of data. 

Since the work of  \cite{paper2, paper4, paper5}, we have constantly pursued the effective learning dynamics, and tried to understand X-form, and more generally, boolean function and boolean circuit. In the process, eventually, we found that the very core of problem is: we need to find a powerful way to describe the property of boolean function. If we have such a tool, we can penetrate into boolean function deep and do much better than before. But, it is not easy to find such a tool. It took us a long time. We recently invented a set of tools, namely, fitting extremum and proper sampling set. Our invention, i.e. fitting extremum and learning dynamics, can be seen in our patent application \cite{patent2, patent3}. How to use fitting extremum and proper sampling set for a spacial case, namely 1-dim real function, can be seen in \cite{paper7}. In this article, we provide theoretical discussions of these tools and related studies. 

We discuss boolean function in section 2, and boolean circuit in section 3. We define a way to present a boolean circuit, i.e. connection matrix, and decomposition of connection matrix. In section 4, we introduce sampling set, fitting extremum, and proper sampling set (PSS). We show the deep connections between PSS and size of boolean circuit. In section 5, we will discuss how to apply these tools to learning dynamics, and prove universal learning machine can be achieved by using them. Finally, in section 6, we make some comments. In appendix, we put details of relationship of PSS and size of boolean circuit.

\section{Boolean Function}
Boolean function and boolean circuit are very important for learning machine. We first define boolean functions and related concepts. 

$\B^N$ is N-dim boolean space, it consists of all N-dim boolean vectors:
$$
\B^N = \{ (b_1, b_2, \ldots b_N) \ \ | \ \  b_k = \text{0 or 1}, \  k = 1, 2, \ldots, N \}
$$

We also called this space as base pattern space \cite{paper2}.  $\B^N$ is the starting point for us.  Specially, when $N = 1, \B^N$  become $\B = \{0, 1\}$. N-dim boolean function is a function defined on $\B^N$:
\begin{definition}[\bf Boolean Function]
A N-dim boolean function $f: \B^N \to \B$ is a function from $\B^N$ to $\B$. We can also write it as: 
$$
f: \B^N \to \B, f(b_1, b_2, \ldots b_N) = \text{0 or 1}
$$
\end{definition} 

We can see some examples of boolean functions.

\begin{exmp}[\bf Some Simplest Boolean Functions]
Constant function is simplest: 
$$
f: \B^N \to \B, \ \ f(b_1, b_2, \ldots b_N) = 1
$$
The function only depends on one variable is also very simple:
$$
f: \B^N \to \B, \ \ f(b_1, b_2, \ldots b_N) = b_1
$$
We can see more examples of boolean function. Boolean functions formed by one basic logic operations are also very simple. Logical operation OR forms one boolean function:
$$
o: \B^2 \to \B, o(b_1, b_2) = b_1 \lor b_2 = \begin{cases} 
      0 & \text{both are 0} \\
      1 & \text{otherwise} 
\end{cases}
$$
Logical operation AND also forms one boolean function:
$$
a: \B^2 \to \B, a(b_1, b_2) = b_1 \land b_2 = \begin{cases} 
      1 & \text{both are 1} \\
      0 & \text{otherwise} 
\end{cases}
$$
Logical operation Identity also forms one boolean function:
$$
id: \B \to \B, id(b)  = b = \begin{cases} 
      1 & b = 1 \\
      0 & b = 0 
\end{cases}
$$
Logical operation Negation also forms one boolean function:
$$
n: \B \to \B, n(b)  = \neg b = \begin{cases} 
      1 & b = 0 \\
      0 & b = 1 
\end{cases}
$$
Logical operation XOR also forms one boolean function:
$$
x: \B^2 \to \B, x(b_1, b_2) = b_1 \oplus b_2 = \begin{cases} 
      1 & \text{one and only of}\ b_1, b_2 \ \text{is zero} \\
      0 & \text{otherwise} 
\end{cases}
$$
It is worth to note that XOR can be written by using OR, AND and Neg: 
$$
b_1 \oplus b_2 = (b_1 \land \neg b_2) \lor (\neg b_1 \land b_2) = (b_1 \lor b_2) \land \neg (b_1 \land b_2) 
$$
\end{exmp}

These simple logic operations are actually form the foundation of boolean function. But boolean functions can be defined and calculated by many ways, not just by logical operations. 

\begin{exmp}[\bf Boolean Function as Real Function]
Logical operation OR can be written as real function:
$$
o: \B^2 \to \B, o(b_1, b_2) = b_1 \lor b_2 = sign(b_1 + b_2), \text{where} \ sign(x) = \begin{cases} 
      1 & x > 0 \\
      0 & x \leq 0 
\end{cases}
$$
where, + is the addition of real number. Logical operation AND can be written:
$$
a: \B^2 \to \B, a(b_1, b_2) = b_1 \land b_2 = b_1 \cdot b_2
$$
where $\cdot$ is the multiplication of real number. Logical operation Negation also forms one boolean function:
$$
n: \B \to \B, n(b) = \neg b = - (b - 1)
$$
\end{exmp}

More boolean function defined by real functions.

\begin{exmp}[\bf More Boolean Functions Defined by Real Functions]
We can define a boolean function as:
$$
f: \B^2 \to \B,  \ \ f(b_1, b_2) = sign(Oscil(r_1b_1 + r_2 b_2)), \ sign(x) =  \begin{cases} 
      1 & x > 0 \\
      0 & x \le 0
\end{cases}
$$
where $r_1, r_2$ are 2 real numbers, sign is the sign function, Oscil is an oscillator function. Oscillator function is something like $sin(x)$, which oscillates from negative to positive and go on. Generally, oscillator functions are very rich. They do not need to be oscillate regularly like $sin(x)$. They could oscillate irregularly and very complicatedly. 

Yet, another boolean function is more popular:
$$
f: \B^N \to \B,  \ \ f(b_1, b_2, \ldots, b_N) = sign(r_1b_1 + r_2 b_2 + \ldots + r_N b_N)
$$
where $r_1, r_2, \ldots, r_N$ are real numbers. This function is often called as a artificial neuron. A little modification will give linear threshold function:
$$
f: \B^N \to \B,  \ \ f(b_1, b_2, \ldots, b_N) = sign(r_1b_1 + r_2 b_2 + \ldots + r_N b_N - \theta)
$$
where $r_1, r_2, \ldots, r_N, \theta$ are real numbers. 
\end{exmp}

Parity function is one important boolean function, which help us in many aspects. 
\begin{exmp}[\bf Parity Function]
Parity function $p: \B^N \to \B$ is defined as below:
$$
p(b_1, b_2, \ldots, b_N) = \begin{cases} 
      1 & \text{number of 1 is odd} \\
      0 & \text{number of 1 is even}
\end{cases}
$$
Parity can also be calculated by real number as below:
$$
p(b_1, b_2, \ldots, b_N) = \Big( \sum_{i = 1}^N b_i \Big) \ (\text{mod 2})
$$
\end{exmp}

Since boolean function is on a finite set, it is possible to express it by a table of value. This table is called as truth table. For example, a parity function of 3 variables can be expressed as below table:
\begin{table}[h]
\centering
    \begin{tabular}{|c|c|c|c|c|c|c|c|c|c|c|c|c|c|c|c|c|}
        \hline
        ~     & $(0,0,0)$ & $(1,0,0)$ & $(0,1,0)$ & $(1,1,0)$ & $(0,0,1)$ & $(1,0,1)$ & $(0,1,1)$ & $(1,1,1)$ \\ 			\hline
       $p(b_1,b_2,b_3)$       & 0     & 1     & 1     & 0     & 1     & 0     & 0     & 1            \\ 
       \hline
    \end{tabular}
\end{table}

We have seen that a boolean function can be defined and calculated by many ways, such as: logical operations, real functions, truth table, etc. But, any boolean function can be expressed by logical operations.

\begin{lemma}[\bf Expressed by Basic Logic Operation]
Any boolean function $f: \B^N \to \B$ can be expressed by basic logic operations: $\lor, \land, \neg$.
\end{lemma} 
{\bf Proof: }First, one boolean function can be expressed by its truth table. In the truth table, there are $2^N$ entries, and at each entry, the function value $f(b_1, b_2, \ldots, b_N)$ is recorded. Since we can use the basic logic operations to express one boolean vector in $\B^N$, each entry can be expressed by basic logic operations. Thus, we can express the boolean function.

For example, we can express the parity function of 3 variables as:
$$
p(b_1, b_2, b_3) = (b_1 \oplus b_2) \oplus b_3
$$
Note, $\oplus$ can be expressed by $\lor, land, \neg$. 

Another example of boolean function.

\begin{exmp}[\bf Expressed By Polynomial Function]
Consider a polynomial function $P$ on real number, e.g., $P(x) = x^3 - 2x^2 - 3$. Also, consider a way to embed a boolean vector $v \in \B^N$ into real number. There are infinite such embeddings. We will consider following:
$$
\forall v \in \B^N, x = b_1 \begin{pmatrix} \dfrac{1}{2}\end{pmatrix} + b_2\begin{pmatrix} \dfrac{1}{2}\end{pmatrix}^2 + \ldots + b_N\begin{pmatrix} \dfrac{1}{2}\end{pmatrix}^N
$$
Then, we define a boolean function $f: \B^N \to \B$ as:
$$
\forall v \in \B^N, f(v) = sign(P(x)), \text{where $x$ is as above}
$$
\end{exmp}

This will define a boolean function on $\B^N$ for any $N$. Such a way to define boolean function and embedding to real number is quite useful. 

\section{Boolean Circuit}
We know a boolean function can be defined and calculated by many possible ways. But, no matter how it is defined and calculated, Lemma 2.1 tells us that it can be expressed by $\lor, \land, \neg$. We call such expression as boolean expression. 

\begin{definition}[\bf Boolean Expression]
A boolean function $f: \B^N \to \B$ can be expressed by $\lor, \land, \neg$ and input variables $b_1, b_2, \ldots, b_N$ as one algebraic expression, we call this algebraic expression as boolean expression of $f$. 
\end{definition} 

Boolean expression is also called boolean formula. 
As one example, the parity function of 4 variable can be expressed as:
$$
p(b_1, b_2, b_3, b_4) = (b_1 \oplus b_2) \oplus (b_3 \oplus b_4)
$$

This is to say, we can realize a boolean function by one algebraic expression. Moreover, we can realize one algebraic expression by hardware that is a group of switches and connections, namely. a circuit. Actually, we can just make such a circuit that is direct translation from the boolean expression, just use a AND switch to replace $\land$, a OR switch to replace $\lor$, and negation connection to replace $\neg$. Thus, we have definition:

\begin{definition}[\bf Boolean Circuit]
Boolean circuit is one directed acyclic graph. There are 2 types of nodes, AND and OR nodes. Connection between nodes are either direct connection (1 to 1 and 0 to 0) or negation connection (1 to 0 and 0 to 1). This graph starts from input nodes: $b_1, b_2, \ldots, b_N$, and ends at the top node. We note that at each node, there are 2 and only 2 connections from below (this is called 2 fanin). But the connections going up could be any number.
\end{definition} 

Note, the definition here are slight different than boolean circuit defined in most literatures (for example \cite{arora+barak}). But, the difference is just very surface and it is just for convenience for our discussions. We can write a boolean circuit in diagram. See diagram below for some examples. A boolean circuit and a boolean expression actually are identical. So, we will later to use them as same.

\begin{exmp}[\bf Some Simple Circuit]
Simplest circuit: $C = 1$. This is a special case. This circuit has no node, i.e. the number of node is 0. 

Second simplest circuit: $C = b_1 \lor b_2$. See Fig. 1 C1 for diagram. This circuit has 1 node and 2 connections. Circuit: $C = b_1 \lor \neg b_2$. See Fig. 1 C3 for diagram. This circuit has 1 node and 2 connections,.one is direct connection, another is negation connection.

Circuit for AND. See Fig. 1 C2 for diagram. This circuit has 1 node and 2 connections, both are direct connections.

Circuit for XOR. See Fig. 1 C5 for diagram. We can express it as: $b_1 \oplus b_2 = (b_1 \land \neg b_2) \lor (\neg b_1 \land b_2)$. This circuit has 3 nodes, i.e. one OR node, and 2 AND nodes, and with 2 negation connection. 

$C = (b_1 \lor (b_2 \land\neg b_3))$. See Fig. 1 C4 for diagram. This circuit has 2 nodes.
\end{exmp}

\begin{center}
\begin{picture}(300,260)(0,0)
\put(-15,0){\resizebox{12 cm}{!}{\includegraphics{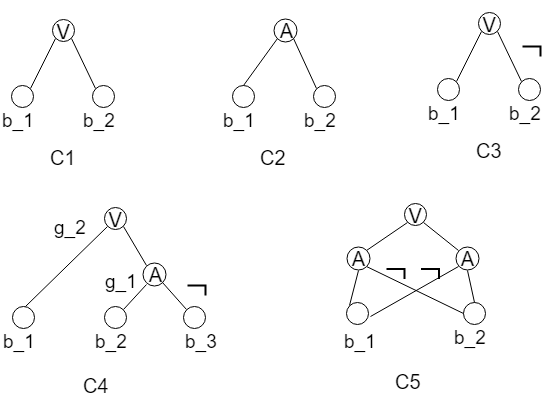}}}
\end{picture}

{\bf Fig1. Diagrams of Some Simple Circuits} 
\end{center}

For a given boolean circuit $C$, for a given input, i.e. $b_1, b_2, \ldots, b_N$ taking value of 0 or 1, we can feed these values into $C$. The circuit will take value at each node accordingly. When the value at the most top node is taken, the circuit take value for itself. This is how a boolean circuit to execute a boolean function. We will denote as $C(b_1, b_2, \ldots, b_N)$.  

Any boolean function $f: \B^N \to \B$, no matter how $f$ is defined and calculated, it can be expressed by one boolean circuit $C$. That is to say, $\forall x \in \B^N, f(x) = C(x)$.     

Clearly, for a boolean function, the boolean circuit to express the function is not unique. For example, one very simple boolean function XOR can be expressed in 2 ways: $(b_1 \land \neg b_2) \lor (\neg b_1 \land b_2)$ or $(b_1 \lor b_2) \land \neg (b_1 \land b_2)$. That is to say, XOR can be expressed by 2 different boolean circuit. For more complicated boolean function, this is even more true.  

A boolean circuit consists of a series of nodes and connections. One very important properties of a boolean circuit is its number of nodes. 

\begin{definition}[\bf Node Number]
For one boolean circuit $C$, we denote the number of nodes of $C$ as $d(C)$. 
\end{definition} 

That is to say, we define a function $d(C)$ on all circuits. Such function is called node number. This function will play an important role in our discussions. 

How can we write a boolean circuit? We can write it as algebraic expression like before. But,  for the purpose of easy manipulation, we need to write them in more ways. First, we denote all nodes of a circuit $C$ as: $g_1, g_2, \ldots, g_d$, where $d = d(C)$. Theses are working nodes. Yet, input variables $b_1, b_2, \ldots, b_N$ are also nodes, which are nodes for inputs. So, $C$ is a graph with nodes $b_1, b_2, \ldots, b_N, g_1, g_2, \ldots, g_d$. $b_1, b_2, \ldots, b_N$ are input nodes, and $g_d$ as ending node, the rest, i.e. $g_i, i =1, 2, \ldots, d$, are working nodes, and $g_d$ is the ending node (it is working node as well). 

At each working node, $g_i, i=1, 2, \ldots, d$, there are 2 and only 2 incoming connections. Except ending node, at each working node, there are 1 or more outgoing connections. 

Thus, besides using diagram and boolean algebraic expression to express a boolean circuit, we can use matrix notation to express a circuit.

\begin{definition}[\bf Connection Matrix]
For a circuit $C$ on $\B^N$, suppose all working nodes of $C$ are $g_1, g_2, \ldots, g_d$, where $d = d(C)$, we define a $d \times (N+d-1)$ matrix $\M$, its entries are these symbols: $\land, \lor, \land\neg, \lor\neg$ or 0, and the meaning of symbols are as following:
$$
\text{at (i, j)}: \begin{cases} 
      0 & \text{no connection from j-th node to i-th working node} \\
      \land & \text{direct connection from j-th node to i-th working node, and this working node is $\land$} \\
      \land\neg & \text{negation connection from j-th node to i-th working node, and this working node is $\land$} \\
      \lor & \text{direct connection from j-th node to i-th working node, and this working node is $\lor$} \\
      \lor\neg & \text{negation connection from j-th node to i-th working node, and this working node is $\lor$}
\end{cases}
$$
We call such maxtrix as connection matrix of $C$. 
\end{definition} 

Clearly, for a given circuit, we can write down its connection matrix. Reversely, if we have such a matrix, it gives a circuit as well. So, we could identify a circuit with a connection matrix.  

We can see some immediate properties of connection matrix. Each row of connection matrix is for one working node, and each column is for connection to all working nodes (except ending node) from one node. Since for each working node, there are 2 and only 2 incoming connections, each row has 2 and only 2 entries are non 0. Since for each node (except ending node), there are 1 or more outgoing connections, each column has 1 or more entries are non 0.  

\begin{exmp}[\bf Examples of Connection Matrix]
Consider a circuit $C_f = b_1 \lor (b_2 \land \neg b_3)$. See Fig. 1 C4 for diagram of this circuit. All nodes of $C_f$ are $b_1, b_2, b_3, g_1, g_2$, and working nodes are $g_1, g_2$, ending node is $g_2$. The connection matrix of $C_f$ is a $2 \times 4$ matrix as below:
$$
\M_f = 
\begin{bmatrix} 
       0    &  \land & \land\neg & 0  \\ 
       \lor    &  0     & 0     & \lor              
\end{bmatrix}
$$
Another example, consider XOR, the circuit is $C_{xor} = (b_1 \lor b_2) \land \neg (b_1 \land b_2)$, all nodes of $C_{xor}$ are $b_1, b_2, g_1, g_2, g_3$, working nodes are $g_1, g_2, g_3$, ending node is $g_3$. The connection matrix of $C_{xor}$ is a $3 \times 4$ matrix as below:
$$
\M_{xor} = 
\begin{bmatrix} 
       \lor    &  \lor & 0 & 0  \\ 
       \land  &  \land   & 0     & 0 \\
       0       &  0    & \land & \land\neg              
\end{bmatrix}
$$ 
\end{exmp}

In the above discussions, there is no order among working nodes. Now we define a order among working nodes. Let's see how the ending node is getting its values. At the very beginning, only input nodes have values, all working nodes are with empty value. When the values propogate along the circuit, the working nodes that have 2 incoming connections from input nodes will get their values. So, these nodes should be put first in the order. But, there could be more than one such nodes. Among these nodes, we will define order by this way: if both 2 nodes $g_i, g_j$ have 2 incoming connections from input nodes, say, $g_i$ with $i_1, i_2, i_1 < i_2$, and $g_j$ with $j_1, j_2, j_1 < j_2$, the order of $g_i, g_j$ are determined by so called dictionary order, i.e. if $i_1 < j_1$, then $g_i$ is first than $g_j$, if $i_1 = j_1, i_2 < j_2$, then $g_i$ is first than $g_j$. Yet, if it is the case: $i_1 = j_1, i_2 = j_2$, then $g_i$ and $g_j$ must be different type (otherwise, we could eliminate one), then the node of $\lor$ is first. 

Now, we have order among working nodes that have 2 incoming connections from input nodes. These nodes will be evaluated. We then consider those working nodes that have 2 incoming connections from nodes that have values already. Then, we can have the order as before. Clearly, we can repeat the above process to give the order to these nodes. So, eventually, we will have the order to all working nodes. 

In one word, the natural order of working nodes means: if one node is evaluated in front, then, it is in front by the natural order. To demonstrate this order, we see one example, circuit for parity function of 4 variables: $C_p = (b_1 \oplus b_2) \oplus (b_3 \oplus b_4)$. See diagram below.

\begin{center}
\begin{picture}(300,180)(0,0)
\put(-15,0){\resizebox{9 cm}{!}{\includegraphics{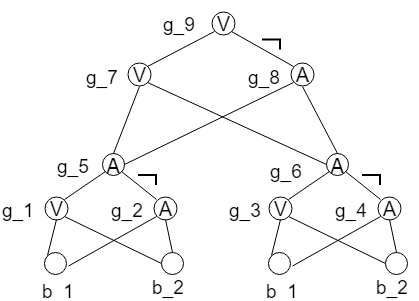}}}
\end{picture}

{\bf Fig2. Circuit of Parity of 4 Variables} 
\end{center}

There are 9 working nodes. Thus, all nodes are $b_1, b_2, b_3, b_4, g_1, g_2, \ldots, g_9$. According to the natural order, working nodes are getting values in this way: $b_1, b_2, b_3, b_4$ get input values, then $g_1, g_2, g_3, g_4$ get values, then, $g_5, g_6$ get values, then, $g_7, g_8$, finally, $g_9$. We can write the connection matrix below. 
$$
\M_p = 
\begin{bmatrix} 
       \lor    &  \lor & 0 & 0  & 0 & 0 & 0 & 0 & 0 & 0 & 0 & 0 \\ 
       \land    &  \land & 0 & 0  & 0 & 0 & 0 & 0 & 0 & 0 & 0 & 0 \\ 
       0    &  0 & \lor & \lor  & 0 & 0 & 0 & 0 & 0 & 0 & 0 & 0 \\ 
       0    &  0 & \land & \land  & 0 & 0 & 0 & 0 & 0 & 0 & 0 & 0 \\ 
       0    &  0    &  0       &  0    & \land & \land\neg & 0 & 0 & 0 & 0 & 0 & 0  \\
       0    &  0    &  0       &  0    &  0       &  0    &  \land & \land\neg    & 0 & 0 & 0 & 0 \\        
       0    &  0 & 0 & 0 & 0 & 0 & 0 & 0 & \lor & \lor  & 0 & 0 \\ 
       0    &  0 & 0 & 0 & 0 & 0 & 0 & 0 &  \land & \land  &0 & 0 \\ 
       0    &  0    &  0       &  0    &  0       &  0    &  0 & 0 & 0 & 0 & \land & \land\neg          
\end{bmatrix}
$$ 

Note, the connection matrix is done according to the natural order of working nodes. If the order in working nodes are different, the connection matrix will appear differently (but just some permutation).

Natural order in working nodes is useful tool. We use a lemma to describe it.

\begin{lemma}[\bf Natural Order of Working Nodes]
For a boolean circuit $C$, suppose its working nodes are $g_1, g_2, \ldots, g_d$, we can make one natural order in the working nodes, so that evaluation of the working nodes will depend on the working nodes in front of it, and will not depend on any working nodes in back of it.
\end{lemma} 
{\bf Proof: }The proof is already done in above discussions. 
\hfill$\blacksquare$

Using the natural order of working nodes, we can see that the working nodes will be in levels. For example, in the example of parity of 4 variables, we have 9 working nodes, and they are divided into 4 levels: level 1: $g_1, g_2, g_3, g_4$, level 2: $g_5, g_6$, level 3: $g_7, g_8$, and level 4: $g_9$. See this clearly in diagram. Nodes in level 1 will get value first. Nodes in level 2, will depends in level 1, etc. That is to say, in order to evaluate nodes in level $j$, all nodes in all levels $i < j$ should be evaluated first.

\begin{definition}[\bf Level of Nodes]
For boolean circuit $C$, suppose its working nodes are $g_1, g_2, \ldots, g_d$, we can group working nodes into a series of subsets $l_1, \ldots, l_K$, $l_i$ consisting of all working nodes that any their incoming connections are from previous subsets, i.e. from $l_j, j < i$. We call each subset $l_i$ as one level of working nodes, we also call the number $K$ as depth, or depth number, or height. 
\end{definition} 

According to Lemma 3.1, we can indeed make such level of working nodes. Clearly, the top level only has one node, i.e. ending node $g_d$. As the above example of parity of 4 variables demonstrates, the evaluation process of a circuit must be level by level. In order to evaluate nodes in level $i+1$, it must first evaluate all nodes in level $i$. This property indicates that we can do decomposition according to level. 

That is to say, we can do evaluation by this way: from input nodes to level 1, then, from level 1 to level 2, etc. If we see the connection matrix of parity of 4 variables, we can see clearly. Thus, we can decompose the connection matrix according to levels. See below:

$$
\M_1 = 
\begin{bmatrix} 
       \lor    &  \lor & 0 & 0   \\ 
       \land    &  \land & 0 & 0   \\ 
       0    &  0 & \lor & \lor  \\ 
       0    &  0 & \land & \land      
\end{bmatrix}
\quad 
\M_2 = 
\begin{bmatrix} 
   \land & \land\neg & 0 & 0   \\
     0    &  0      &  \land & \land\neg      
\end{bmatrix}
\quad 
\M_3 = 
\begin{bmatrix} 
       \lor    &  \lor   \\ 
       \land    &  \land     
\end{bmatrix}
\quad 
\M_4 = 
\begin{bmatrix} 
     \land & \land\neg 
\end{bmatrix}
$$ 
Here, $\M_1$ is for: from input nodes to get value of nodes in level 1. For example, if $v = (1, 0, 0, 1)^T$ is the input, then $\M_1 v = (1, 0, 1, 0)^T$. vecto r$(1, 0, 1, 0)^T$ gives values of all nodes in level 1. We can continue to use $\M_2$ for values of all nodes in level 2, $\M_3$ for values of all nodes in level 3, and finally, $\M_4$ for value of top node. We can write these operations into following form:
$$
C_p(v) = \M_4 \M_3 \M_2 \M_1 v, \quad v = (b_1, b_2, b_3, b_4)^T \in \B^4
$$

Here, $C_p$ is the circuit of parity of 4 variables, and $C_p(v)$ stands for the value of top node, which is the output value of the circuit. In this way, we can operate on circuit much easier. It is still not as good as ordinary matrix calculations, but it is much better and clear. We will use this notation consistently. 

However, we need to be more careful. In the above example, level $i+1$ only depends on level $i$, not on level $i-1$ directly. This is not always true. Consider the circuit $C_f$, which is in diagram of C4 in Fig. 1. All nodes of $C_f$ are $b_1, b_2, b_3, g_1, g_2$. Working nodes are $g_1, g_2$. The connection matrix of $C_f$ is a $2 \times 4$ matrix as below:
$$
\M_f = 
\begin{bmatrix} 
       0    &  \land & \land\neg & 0  \\ 
       \lor    &  0     & 0     & \lor              
\end{bmatrix}
$$
So, clearly, level 0 is \{$b_1, b_2, b_3$\} (input nodes), level 1 is \{$g_1$\}, level 2 is \{$g_2$\} (ending node). But, we can see that level 2 node has incoming connections from level 1 and level 0. Thus, the decomposition according to level to level seems has difficulties. Can we still do decomposition as we did for $C_p$?

In order to make neat decomposition, we need introduce a new kind of node: spurious node. We will se $s$ to denote spurious node. A spurious node is one node adding to one level to just pass the connections from lower level to higher level. After introducing spurious node, then, we can go back to the situation: level $i+1$ will only depends on level $i$, not on any previous level. As one example to demonstrate, for $C_f$, we add one spurious node in level 1. This spurious node has 1 and only 1 incoming connection, and this node will not do anything, but pass the value of $b_1$, so its outgoing connections are exactly same as the outgoing connections of $b_1$. So, after add this node,  $g_2$ will have 2 incoming connections from level 1. So, we can write following decomposition.  
$$
\M_1 = 
\begin{bmatrix} 
       s   &  0 & 0 \\ 
       0   &   \land & \land\neg    \\ 
\end{bmatrix}
\quad 
\M_2 = 
\begin{bmatrix} 
     \lor & \lor      
\end{bmatrix}
\quad 
$$ 
And,
$$
C_f(v) = \M_2 \M_1 v, \quad v = (b_1, b_2, b_3)^T \in \B^3
$$

This decomposition will make our operation on circuit easier. For example, if input is $v = (1, 1, 0)^T$, then, $u = \M_1 v = (1, 1)^T$, $\M_2 u = 1$, so $C_f(v) = 1$. 
 
\begin{definition}[\bf Spurious Nodes]
For a circuit $C$ on $\B^N$, suppose all working nodes of $C$ are $g_1, g_2, \ldots, g_d$, where $d = d(C)$, and nodes are grouped into levels: $\{g_{i, j}, i=1, \ldots, K, j=1,\ldots,L_i$, where $K$ is the number of levels. If at level $i+1$, there are the incoming connections not from level $i$, but from level lower than $i$, we can add spurious nodes in level $i$, so that these nodes only pass the value. We use $s$ to denote such nodes. By adding spurious nodes, the evaluation of one level $i+1$ will only depend on level $i$.  
\end{definition} 

We can write this decomposition into following lemma.

\begin{lemma}[\bf Decomposition of Connection Matrix by Level]
For a boolean circut $C$, suppose its working nodes are $g_1, g_2, \ldots, g_d$,  and nodes are grouped into levels: $\{g_{i, j}, i=1, \ldots, K, j=1,\ldots,L_i$, where $K$ is the number of levels. Then, if necessary, we can add spurious nodes, then the evaluation of $C$ will be decomposited to a series evaluation so that each evaluation is done from one level to next level. And, each evaluation can be achieved by matrix operation. 
\end{lemma} 
{\bf Proof: }The proof is already done in above discussions. 
\hfill$\blacksquare$

\begin{center}
\begin{picture}(350,220)(0,0)
\put(-15,0){\resizebox{14 cm}{!}{\includegraphics{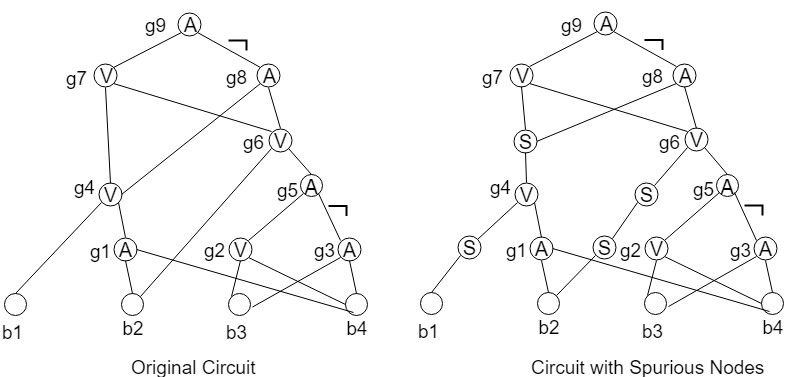}}}
\end{picture}

{\bf Fig. 3 Circuit of 5 Levels} 
\end{center}

\begin{exmp}[\bf Example of Decomposition]
We consider this boolean circuit: $C = (b_1 \lor (b_2 \land b_4)) \oplus ( b_2 \lor (b_3 \oplus b_4))$. See the diagram for this circuit in Fig. 3, which is the left diagram. $C$ has 9 working nodes: $g_1, g_2, \ldots, g_9$. The working nodes are ordered as we discussed before. We can write down working nodes as: $g_1: b_2 \land b_4$, $g_2: b_3 \lor b_4$,  $g3: b_3 \land b_4$, $g_4: b_1 \lor g_1$,  $g_5: g_2 \land\neg g_3$, $g_6: b_2 \lor g_5$, $g_7: g_4 \lor g_6$, $g_8: g_4 \lor g_6$, $g_9: g_7 \land\neg g_8$.
The connection matrix is blow:
$$
\M_p = 
\begin{bmatrix} 
       0    &  \land & 0 & \land  & 0 & 0 & 0 & 0 & 0 & 0 & 0 & 0 \\ 
       0    &  0 & \lor & \lor  & 0 & 0 & 0 & 0 & 0 & 0 & 0 & 0 \\ 
       0    &  0 & \land & \land  & 0 & 0 & 0 & 0 & 0 & 0 & 0 & 0 \\ 
       \lor   &  0    &  0       &  0    & \lor & 0 & 0 & 0 & 0 & 0 & 0 & 0  \\
       0    &  0  & 0 & 0  & 0 & \land & \land\neg & 0 & 0 & 0 & 0 & 0 \\ 
       0    &  \lor    &  0       &  0    &  0       &  0    &   0    & \lor   & 0 & 0 & 0 & 0 \\        
       0    &  0 & 0 & 0 & 0 & 0 & 0 & 0 & \lor & \lor  & 0 & 0 \\ 
       0    &  0 & 0 & 0 & 0 & 0 & 0 & 0 &  \land & \land  &0 & 0 \\ 
       0    &  0    &  0       &  0    &  0       &  0    &  0 & 0 & 0 & 0 & \land & \land\neg          
\end{bmatrix}
$$ 
There are 5 levels in this circuit: level 0: \{$b_1, b_2, b_3, b_4$\}, level 1: \{$g_1, g_2, g_3$\}, level 2: \{$g_4, g_5$\}, level 3:  \{$g_6$\}, level 4:  \{$g_7, g_8$\}, level 5:  \{$g_9$\}. These levels are not single level evaulation. For example, at $g_4$, we need $g_1$ (level 1) and $b_1$ (level 0) to evaluate it. But, we can add spurious nodes. See the right diagram in Fig. 3, where nodes $S$ are spurious nodes. We can see clearly, with spurious nodes, the circuit becomes single level evaluation. Then, we can do decomposition by level. We have following connection matries between levels.
$$
\M_1 = 
\begin{bmatrix} 
      s & 0 & 0 & 0  \\      
      0 & \land & 0 & \land  \\ 
       0 & s & 0 & 0     \\
       0 & 0 & \lor & \lor  \\ 
       0 & 0 & \land & \land
\end{bmatrix}
\quad 
\M_2 = 
\begin{bmatrix} 
       \lor   & \lor & 0 & 0  & 0  \\
      0 & 0 & s & 0 & 0      \\
       0    &   0 & 0  & \land & \land\neg
\end{bmatrix}
\quad 
\M_3 = 
\begin{bmatrix} 
     s & 0 & 0 \\     
   0   &  \land & \land\neg   \\
\end{bmatrix}
$$
$$
\M_4 = 
\begin{bmatrix} 
       \lor & \lor \\
       \land & \land
\end{bmatrix}
\quad 
\M_5 = 
\begin{bmatrix} 
      \land & \land\neg 
\end{bmatrix}
$$ 
By using these connection matrices, we can see the evaluation of circuit as following: 
$$
C(v) = \M_5\M_4\M_3\M_2 \M_1 v, \quad v = (b_1, b_2, b_3, b_4)^T \in \B^4
$$
First, input value is $v = (b_1, b_2, b_3, b_4)^T$. We feed this into $M_1$, and get a 5-dim vector $u = \M_1v$. Then, feed $u$ into $\M_2$, we will get a 3-dim vector $u' = \M_2u$. Then, feed into $\M_3$, get a 2-dim vector. Then, feed into $\M_4$, get a 2-dim vetor. Finally, feed into $\M_5$, get the value at ending node.
\end{exmp}
Note, the role that spurious nodes are playing. 

This example shows that decomposition will make boolean circuit becomes much easier to analyze. After decomposition, we have several levels. Each level is very simple boolean circuits: each node has only 2 incoming connections, and all nodes are in the exactly same level. We can use one matrix to record this one level circuit well. We call this matrix as one level connection matrix. We can use the matrix to evaluate all nodes in the one level circuit, and the evaluation is very simple and mechanical, almost like the normal matrix-vector multiplication. This makes analysis much easier. Although the operation is not truly matrix calculation, yet, it is quite simple and easier to handle. So, the above notation is good enough to help us to record the boolean circuits, and help us the do operations and analysis on boolean circuits.

\subsection*{About Size of Boolean Circuit}
In most literatures about boolean circuit, for example, in \cite{arora+barak}, the size and depth of a boolean circuit are defined. They are highly related to and different from our definition of node number and level number. We discuss them here. 

In \cite{arora+barak}, the size of a boolean circuit is defined as the number of gates $\lor, \land, \neg$ used in the circuit. In contrast, we define the node number of a circuit as the number of nodes $\lor, \land$, not including $\neg$. We will use notation $s(C)$ for size of a boolean circuit (as most literature), and use notation $d(C)$ for node number. 

In most literatures, the depth of circuit is defined as: the steps required from input to output. Our definition of depth is exactly same as most literature. The depth equals the number of levels. So, if depth of a circuit is $K$, we can decompose connection matrix to $K$ connection matrices, and each such matrix is only for one level, i.e. depth is 1.  

\begin{lemma}[\bf Relationship of $s(C)$ and $d(C)$, $K$ and Depth]
For a boolean circuit $C$, suppose $s(C)$ is the size of circuit (as most literature), and $d(C)$ is node number, then $d(C) \le s(C) \le 3d(C)$. And, depth of a circuit equals number of levels.
\end{lemma} 
{\bf Proof: }The proof is clear. 
\hfill$\blacksquare$

Since circuit complexity in most literature is measured by $s(C)$, if we are interested in circuit complexity, using $d(C)$ is equivalent to using $s(C)$. However, for our purpose, to use $d(C)$ is more convenient. We will mostly use $d(C)$ to measure a circuit.

\section{Fitting Extremum and Proper Sampling Set}
In order to analyze boolean function $f: \B^N \to \B$, one way is to consider some examples, say, we feed some $x \in \B^N$ into $f$ and see its value. This is called sampling. More precisely, we get an input $x \in \B^N$ by some way, we then get value of $f(x)$, this forms one sample of $f$. If we repeat such sampling for some times, we get the sampling set. 

\begin{definition}[\bf Sampling Set]
A sampling set is one subset of $\B^N$, that is, if $S \subset \B^N$, we say $S$ is one sampling set (or, just sampling). Moreover, over one sampling set, there could have assigned values:
$$
Sv = \{ [x, b] \ | \ x \in S, \ b = \text{0 or 1}  \}
$$
We say such set $Sv$ as sampling set with assigned values, or sampling with values, or just sampling. For a boolean function $f: \B^N \to \B$,  we can have the sampling set of $f$ (or sampling set for $f$): 
$$
Sv = \{ [x, f(x)] \ | \ x \in S \}
$$
\end{definition} 

Sampling set of $f$ will give us information about this boolean function. We can think a sampling set of $f$ as a subset of the truth table of $f$. Naturally, we want to ask: Can we recover the whole truth table by a sampling set? Actually, under certain condition, we can. See this simple example. Consider the simplest circuit: $C = b_1 \lor b_2$. The truth table is very simple as below:
\begin{table}[h]
\centering
    \begin{tabular}{|c|c|c|c|c|c|c|c|c|c|c|c|c|c|c|c|c|}
        \hline
        ~     & $(0,0)$ & $(1,0)$ & $(0,1)$ & $(1,1)$  \\ \hline
       $C(b_1,b_2)$       & 0     & 1     & 1     & 1     \\ 
        \hline
    \end{tabular}
\end{table}
If we only have a subset of this truth table, can we use a it to recover the whole truth table? Depends. If the subset is: $[(0, 0), 0], [(1, 0), 1]$, we could not, since there is another circuit $C = b_1$ satisfies this sampling set as well. But, if the subset is:  $[(0, 0), 0], [(1, 0), 1], [(0, 1), 1]$, we could. Even though this subset is a true subset of truth table, we can see clearly, there is no any other {\it simple} circuit satisfies this set. But, there is indeed a circuit $C_{xor}$ satisfies this set and it is not $C$. But, this circuit $C_{xor}$ is more complicated than $C$, i.e. it has more nodes. 

This simple fact, of course many other facts as well, motivates us to consider this question: Given a sampling set, if we look a simplest boolean circuit to satisfy the sampling set, what would happen? Can we recover the whole truth table by this action? This is the central question that we try to address. But first we define circuit space.

\begin{definition}[\bf Circuit Space on $\B^N$]
The set of all boolean circuit on $\B^N$ is called circuit space on $\B^N$. We use $\C$ to represent the circuit space.  
$$
\C = \{ C \ | \ C \text{ is boolean circuit on } \B^N\}
$$
\end{definition} 

Note, $\C$ is much a bigger set than the set of all boolean functions. The number of boolean functions on $\B^N$ are finite, though the number is very huge: $2^{2^N}$. But, one boolean function could have many boolean circuits to express it. So, $\C$ is a much larger space. 

We then define Fitting Extremum that is a minimizing problem to look for the boolean circuit that has smallest node number while fitting with sampling.

\begin{definition}[\bf Fitting Extremum]
For a sampling set $Sv$ with values, we define one extremum problem as following:
$$
\text{Min:}\ d(C), \ C \in \C  \ \& \  \forall [x, b] \in Sv \ C(x) = b 
$$
We call this problem as fitting extremum on $Sv$.
\end{definition} 

In fitting extremum, we are looking for boolean circuit in $\C$ that it has these properties: 1) fitting with sampling set and 2) with smallest node number. We can use one most simple case to illustrate the meaning of fitting extremum. Consider sampling set: $\{[(0, 0), 0], [(1, 0), 1], [(0, 1), 1]\}$. As discussed above, this could be a subset of truth table of some unknown circuit. We want to use this sampling set to recover the whole truth table. When we look circuit fitting with sampling, we find that 2 circuits $C_1 = b_1 \lor b_2$ and $C_2 = b_1 \oplus b_2$ fitting with sampling. So, which circuit should we choose? Just sampling set itself is not good enough. But, if we add one more condition, i.e. to look for simplest circuit fitting with sampling, then, we know $C_1$ should be chosen, since $d(C_1) = 1, d(C_2) = 3$. This simple example indeed tells us what fitting extremum is about.   

In the definition of fitting extremum, we give a sampling set with values. But, what if we give a subset of $\B^N$ and a boolean function? This sure will give a fitting extremum as well.  

\begin{definition}[\bf Fitting Extremum of a Boolean Function]
For one boolean function $f: \B^N \to \B$, and for a sampling set $S \subset \B^N$, we define one extremum problem as following:
$$
\text{Min:}\ d(C), \ C \in \C  \ \& \  \forall x \in S \ C(x) = f(x) 
$$
We call this problem as fitting extremum on $S$ and $f$.
\end{definition}   

Such a circuit $C$ is called as circuit generated by fitting extremum on sampling $S$ and $f$. That is to say, given a sampling and a boolean function, we can generate a circuits from them. 

\begin{lemma}[\bf Existence of Circuit Generated]
For any given boolean function $f$, and any given sampling $S$, the circuit generate by fitting extremum on  $S$ and $f$ always exists. That is to say, there exists at least one circuit $C$ so that $C$ fitting with sampling and $d(C)$ reach minimum. 
\end{lemma}
{\bf Proof:} For a given $S$, we denote the set of circuits as $G$: $G = \{C \in \C \ | \ C \text{ fits with with } S\}$. Very clear that $G$ is not empty, since there is  at least a circuit $C$ expressing $f$, then $C$ fits with $S$. So, the set $\{d(C) \ | \ C \in G\}$ is a nonempty set of integers. Thus, there must be a $C$ so that $d(C)$ equals the minimum.
\hfill$\blacksquare$

So, for any given $f$ and $S$, there is at least one circuit $C$ generated by fitting extremum from them. That is to say, if we have a boolean function $f$ and a sampling set $S$, we can put them into fitting extremum, then we get one or more boolean circuit $C$ fitting with $f$ on $S$. Naturally, we ask: what is the relationship between $C$ and $f$? Could this circuit $C$ express $f$ exactly? We first see a simple example.

For OR function $f = b_1 \lor b_2$, for sampling  $S = \{(1, 0), (0, 0)\}$, if we put them into fitting extremum, it is easy to see circuit $C = b_1$ fitting with sampling and $d(C) = 0$. So, circuit $b_1$ is a circuit generated by fitting extreme. But, the circuit $C$ does not express $f$ since $C(0, 1) \neq f(0, 1)$. However, if we choose sampling $S =  \{(1, 0),(0, 1), (0, 0)\}$, the circuit generated by fitting extremum from $f$ and $S$ is $C = b_1 \lor b_2$, which expresses $f$ exactly. 

This simple example tells us: For a boolean function $f$, for some sampling $S$, the circuit $C$ generated by fitting extremum from $f$ and $S$ indeed expresses $f$, but for some other sampling, the circuit generated from fitting extremum does not express $f$. The sampling that makes fitting extremum to produce a circuit expressing $f$ is special and needs our attention. Thus, we define proper sampling set.   

\begin{definition}[\bf Proper Sampling Set]
For a given boolean function $f: \B^N \to \B$, and for a sampling set $S \subset \B^N$, if fitting extremum on $S$ and $f$ generates a boolean circuit $C$, i.e. $C$ fits $f$ on $S$, and $d(C)$ reaches minimum, and if $C$ expresses $f$ exactly, i.e. $\forall x \in \B^N, C(x) = f(x)$, we say $S$ is a proper sampling set of $f$, or just proper sampling. 
\end{definition} 

In another words, when $S$ is proper sampling set, the boolean circuit generated by fitting extremum on $S$ and $f$ will always express $f$. This is one crucial property. 

We will use PSS to stand for proper sampling set. In the above simple example, for OR function $f$, $S_1 = \{(1, 0), (0, 0)\}$ is not PSS, but $S_2 =  \{(1, 0),(0, 1), (0, 0)\}$ is PSS.

\begin{lemma}[\bf Existence of PSS]
For any boolean function $f$, there is some subset $S \subset \B^N$ so that $S$ is proper sampling set of $f$. 
\end{lemma}
{\bf Proof:} This is very clear. At least, the whole space $\B^N$ is proper sampling. 
\hfill$\blacksquare$

That is to say, for any boolean function $f$, PSS always exists. The trivial case is that PSS equals the whole boolean space $\B^N$. We can think in this way: give a sampling $S$, if $S$ is not PSS, we can add more elements into $S$, eventually, $S$ will become PSS. Of course, we do not want the whole space, if possible. This is actually the major problem we will discuss here. First, we consider more examples.

\begin{exmp}[\bf Examples for Sampling and PSS]
Note, normally, we write vectors as column. But, for convenience, for short vectors (low dimension), we write as row.

For OR function $f = b_1 \lor b_2$, the sample set $\{(1, 1)\}$ is not PSS. It is easy to see the fitting extremum generate a constant circuit $C = 1$. But, the sampling set $S = \{(1, 0), (0, 1), (0, 0)\}$ is PSS. Fitting extremum generates $C = b_1 \lor b_2$, which expresses $f$ exactly. Note, $|S| = 3$.  

For AND function $f = b_1 \land b_2$, the sampling set $\{(0, 1), (1, 1)\}$ is not PSS. It is easy to see, fitting extremum generates a circuit $C = b_2$. But, the sampling set $S = \{(1, 1), (0, 1), (1, 0)\}$ is PSS, fitting extremum generates $C = b_1 \land b_2$, which expresses $f$. Also note $|S|=3$.

For XOR function $C = b_1 \oplus b_2$, the sampling set $S = \{(1, 0), (0, 1)\}$ is not PSS. But, \\ $S = \{(1, 0), (0, 1), (1, 1), (0, 0)\}$ is PSS. Here, $|S| = 4$.  

See diagram C4 in Fig. 1. It is for a function $f = b_1 \lor (b_2 \land\neg b_3)$. Sampling \\$S = \{(1,0,0), (0,1,0), (0,1,1), (0,0,0)\}$ is PSS. How do we know this? Let's see some details. For node $g_1 = b2 \land\neg b_3$, this is a $\land$ node with one negation connection. As we talked above, for $\land$ node, the PSS should be: $\{(1, 1), (0, 1), (1, 0)\}$, but, since there is  one negation connection, for $\land\neg$ node, the PSS become: $\{(1, 0), (0, 0), (1, 1)\}$. This is only for $b_2, b_3$. But, we can add $b_1$ as 0, so, we have a set $\{(0,1,0), (0,0,0), (0,1,1)\}$. But, we need sampling for $b_1$. This is the sampling $(1,0,0)$, as we set $b_1$ as 1, and $b_2, b_3$ as 0. So, we have $S = \{(1,0,0), (0,1,0), (0,1,1), (0,0,0)\}$. We then consider node $g_2 = b_1 \lor g_1$. This is $\lor$ node. As above discussion, for this node, we need to have $\{(1,0), (0,1), (0,0)\}$ for $b_1, g_1$. But, for this case, $S$ indeed will cause to have $\{(1,0), (0,1), (0,0)\}$ for $b_1, g_1$. Thus, $S$ is a PSS. We can verify this by trying some circuits. But, the procedure we did here is generally true, which we will see in later discussions.
\end{exmp}

\begin{exmp}[\bf More example of PSS]
Consider a sampling with value, in $\B^2$, $Sv = \{[(1,1), 1], [(0,0), 0]\}$. This sampling set is not PSS. We can easily see that circuit $C_1 = b_1$ fits with $S$, and $C_2 = b_2$ fits with $S$ as well. However, if we add one more sampling into $S$, for example: $[(1,0), 1]$, we can exclude out $C_2 = b_2$. Thus, $S = \{[(1,1), 1], [(0,0), 0], [(1,0), 1]\}$ is a PSS. 
\end{exmp}

From above discussions, we know that for a boolean function $f$, we could first sampling it, then apply fitting extremum on sampling, if the sampling is right, i.e. it is PSS, we will get a boolean circuit that express $f$. This is a very great outcome. With this procedure, we can understand $f$ better.

\begin{theorem}[\bf PSS implies Circuit]
If $f$ is a boolean function $f: \B^N \to \B$, and $S \subset \B^N$ is a PSS for $f$, and $|S|$ is the size of PSS, then there is a circuit $C$ expresses $f$ and $d(C) < N|S|$. 
\end{theorem}

Opposite direction is also true, that is to say, if we have circuit, we can to construct a PSS from it. 

\begin{theorem}[\bf Circuit implies PSS]
If $f$ is a boolean function $f: \B^N \to \B$, and $C$ is a boolean circuit to express $f$, then there is a PSS for $f$, and size of PSS is less than $3d(C)$. 
\end{theorem}

PSS implies circuit theorem tells us that for a boolean function $f$, if we have a PSS for $f$, we can construct a circuit to express $f$ and the size of circuit is controlled by size of PSS. Note, the size of circuit is one good measure of complexity of $f$, thus, the size of PSS is also a good measure of complexity of $f$. 

Circuit implies PSS theorem tells us that for a boolean function $f$, if we know a circuit $C$ expressing $f$, we can pick up PSS by using $C$. 
 
So, the 2 theorems tell us that for a boolean function $f$, if we have a PSS of $f$, we can construct a circuit to express $f$ and the size of circuit is controlled by size of sampling. And, reversely, if there is one circuit expressing $f$, then we can find a PSS by using circuit, and the size of sampling is controlled by size of circuit. Thus, the size of circuit and size of PSS is equivalent. Since the size of circuit is one good measure of computational complexity of $f$, so is the size of PSS. This is a very important property.

The above 2 theorems are very crucial. We put the proofs for them in Appendix.

For one boolean function $f$, there might be more than one PSS of it. Could be many. But, among all PSSs, the PSS with lowest number of nodes will be specially interesting.

\begin{definition}[\bf Minimal Proper Sampling Set]
For a given boolean function $f: \B^N \to \B$, if a sampling $S \subset \B^N$ is a proper sampling set, and $|S|$ reaches the minimum, we call such a sampling set as minimal proper sampling set. 
\end{definition} 

We use brief notation mPSS for minimal proper sampling set.

\section{Learning Dynamics}
We discussed universal learning machine in \cite{paper2, paper4, paper5}, which is a machine that can learn any possible to learn without human intervention. In our previous discussions, the learning dynamics of universal learning machine was given special attention, and several methods/strategies were introduced. As the result, we proved that with sufficient data (sufficient to bound and sufficient to support), universal learning machine can be realized. Of course, we are constantly looking for better learning methods. As a matter of fact, we invented Fitting Extremum and Proper Sampling Set (FE and PSS) particularly for such a purpose. Without the efforts to find better learning methods, perhaps FE and PSS would not be invented. In this section, we will discuss on how to utilize FE and PSS for learning dynamics.

\subsection*{Universal Learning Machine}
We briefly recall learning machine and learning dynamics. 
An universal learning machine $\mm$ is a system consisting of input space, output space, conceiving space and governing space. The input space has $N$ dimension, and output space has $M$ dimension. The conceiving space contains information processing unit that will get information from input space, process the information, and put results into output space. The conceiving space is the container for information processing units, and it normally contains many pieces of information processing. But, at one particular time, only one information processing unit is used to generate output. The learning is actually modifying/adapting the current information processing unit so that it becomes better. Governing space is the container for methods that control how learning is conducted.

For convenience of discussions and without loss of generality, we often set the dimension of output space $M=1$. Thus the information processing unit becomes a boolean function $p: \B^N \to \B$. Inside conceiving space, there could be many boolean functions, and one is used as current information processing unit. 

The input space is $N$ dimension, thus input $v \in \B^N$. We also call the space $\B^N$ as {\it base pattern space}. Any vector $v \in \B^N$ is also called as a base pattern. Learning machine will get information from input $v$ and form subjective view for $v$ in machine. Such subjective view is called as subjective pattern, which is handled inside machine by something called X-form. Actually, the information processing is done according to those subjective patterns, so according to X-forms. Inside conceiving space, normally, there are many X-forms.   

X-form plays one crucial role in learning machine. For full details of X-form, consult \cite{paper2, paper4, paper5}. Here, we focus on relationship between X-form and boolean functions.

\begin{definition}[\bf X-form as Algebraic Expression]
If $E$ is an algebraic expression of 3 operators, $\lor, \land, \neg$ (OR, AND, NOT), and $g = \{b_1, b_2, \ldots, b_K\}$ is a group of base patterns, then we call the expression $E(g) = E(b_1, b_2, \ldots, b_K)$ as an X-form upon $g$, or simply X-form. 
\end{definition}

Note a small difference on surface: in \cite{paper2, paper4, paper5}, we used $+, \cdot, \neg$ for OR, AND, NOT operators. In fact, if we want to do algebraic expression, to use $+, \cdot, \neg$ is much better. Here, for consistence with this paper, we use $\lor, \land, \neg$, though, which is not as good for algebraic expressions.

In another words, a X-form is an algebraic expression of some base patterns. This is one way to see X-form. But, we can view such algebraic expression as subjective pattern.

\begin{definition}[X-form as Subjective Pattern]
Suppose $g =\{p_1, p_2, \ldots, p_K\}$ is a set of subjective pattern, and $E = E(g ) = E(p_1, p_2, \ldots, p_K)$ is one X-form on $g$ (as algebraic expression). With necessary supports (i.e. the operations in the algebraic expression can be realized), this expression $E$ is a new subjective pattern.
\end{definition}

Further, such algebraic expression can be viewed as information processing:

\begin{definition}[X-form as Information Processor]
Assuming $\mm$ is a learning machine, $g =\{p_1, p_2, \ldots, p_K\}$ is a set of subjective patterns subjectively perceived by $\mm$, and $E = E(g)$ is a X-form on $g$ (as algebraic expression), then $E(g)$ is an information processing unit that processes information like this: when a basic pattern $p \in \B^N$ is put into $\mm$, and $\mm$ perceives this pattern, then the subjective patterns $p_1, p_2, \ldots, p_K$ forms a set of boolean variables, still written as: $p_1, p_2, \ldots, p_K$, and when this set of boolean variables is applied to $E$, the value of $E$ is the output of the unit, and it is written as: $E(g)(p)$.
\end{definition}

Thus, one X-form actually is one boolean function. So, we now understand the meaning of X-form in several aspects. Why do we call as X-form? These expressions are mathematical forms and have very rich meanings, yet there are many properties of such expressions are unknown. Following tradition, we use X to name it. 

Following theorem connect objective pattern, subjective pattern and X-form. 

\begin{theorem}[Objective and Subjective Pattern, and X-form]
Suppose $\mm$ is an learning machine. For any objective pattern $p_o$ (i.e. a subset in $\B^N$), we can find a set of base pattern $g =\{b_1, b_2, \ldots, b_K\}$, and one X-form $E$ on $g$, $E = E(g) = E(b_1, b_2, \ldots, b_K)$, so that $\mm$ perceives any base pattern in $p_o$ as $E$, and we write as $p_o = E(g)$. We say $p_o$ is expressed by X-form $E(g)$.
\end{theorem}

We skip the proof here, which can be found in \cite{paper2}.

\begin{exmp}[\bf X-form and related] We see some examples for X-form.\\
{\bf 1: }Suppose $N = 2$ and the information processing unit is such a boolean function: $f: \B^2 \to \B, f(b_1, b_2) = b_1 \oplus b_2$. We can write this boolean function in X-form. Let $p_1 = (1, 0), p_2 = (0, 1)$, so $p_1, p_2$ both are base patterns, and one algebraic expression $E(p_1, p_2) = (p_1 \lor p_2) \land \neg (p_1 \land p_2)$, then we can see: for any $v \in \B^2, E(v) = f(v)$. 

{\bf 2: }Suppose $N = 3$, we have one objective pattern $p_o = \{(0,0,0), (1,0,0), (1,1,0), (0,1,0)\}$, we can have these base patterns: $\{p_1, p_2, p_3\}, p_1 = (1,0,0), p_2 = (0,1,0), p_3 = (0,0,0)$, and algebraic expression $E(p_1, p_2, p_3) = p_3 \lor p_1 \lor p_2 \lor (p_1 \land p_2)$, so that $p_o = E(p_1, p_2, p_3)$. We can see the number of operations in $E$ is $d(E)=4$.   

{\bf 3: }Suppose $N = 4$, we have some X-forms: $Q_1, Q_2, Q_3$, then, we can form new X-forms as: $(Q_1 \lor Q_2) \land \neg (Q_2 \land Q_3)$
\end{exmp}

If we want to emphasis the information processing unit, we can just focus on boolean function. But, in this way, we lost the connection to subjective pattern that is crucial in many aspects. By using X-forms. we can reach both subjective pattern and boolean function, since X-form is both. Inside conceiving space, there are a lot of X-forms. We can find some X-forms are better, and choose them. And, we use existing X-forms to form new X-form that would be better. These actions are actually learning dynamics. Following learning strategies will tell us how to do learning. 

\begin{lemma}
If $E$ is a X-form, then there is a boolean circuit $C$, so that $\forall p \in \B^N$, $E(p) = C(p)$. and $d(C) = d(E) + L$, where $d(C)$ is the number of nodes of $C$, $d(E)$ is the number of operators $\land$ and $\lor$ in $E$, $L$ is an adjusting number. 
\end{lemma}
{\bf Proof:}  $E$ is an X-form, according to definition, there is an algebraic expression of 3 operators, $\lor, \land, \neg$ (OR, AND, NOT), and $g = \{p_1, p_2, \ldots, p_K\}$ is a group of base patterns, and $E = E(b_1, b_2, \ldots, b_K)$. Note, $E$ is almost a boolean circuit, there are only 2 things are different. One is: in $E$, there are 3 operators, $\neg$ is view as one operator. But, in boolean circuit $C$, $\neg$ is integrated into node. So, if we only count $\lor$ and $\land$ operators in $E$, we can get the number of nodes of $C$. Another difference is: $E$ is based on base patterns: $\{p_1, p_2, \ldots, p_K\}$. But, we write base patterns $p_1, p_2, \ldots$ into the form: $p_1 =  (\ldots(s_1 b_1 \land s_2 b_2) \land s_3 b_3) \ldots \land s_N b_N)$, where $s_i, i=1,2,\ldots,N$ are: if $b_i = 1, s_i = id$, if $b_i = 0, s_i = \neg$. We can do same for $p_2$, etc. (see the Lemma 4.3 circuit of a single vector). We connect these circuits with $E$, we then have the boolean circuit $C$ that expresses the X-form $E$. Also, $d(C) = d(E) + L$, where $L$ depends on 1) the number of $\neg$ in $E$, 2) the number of nodes used in $p_1$ etc, which is $K(N-1)$.
\hfill$\blacksquare$

This lemma tells us that we can get a boolean circuit from a X-form. Reversely, we can also get one X-form from a boolean circuit.  

\begin{lemma}
If $C$ is a boolean circuit over $\B^N$, it is an X-form $E$ as well, and $d(C) = d(E) + L$, where $d(C)$ is the number of nodes of $C$, $d(E)$ is the number of operators $\land$ and $\lor$ in $E$, $L$ is an adjusting number.
\end{lemma}
{\bf Proof:} $C$ is a boolean circuit, so it is such: there is an algebraic expression $E$ of 3 operators, $\lor, \land, \neg$ (OR, AND, NOT), and this expression $E$ on this group of base pattern: $g = \{b_1, b_2, \ldots, b_N\}$, the $C = E(b_1, b_2, \ldots, b_N)$. Clearly, $E$ is an X-form. We also see $d(C) = d(E) +L$. 
\hfill$\blacksquare$

We point out here: $C$ is a boolean circuit that is objective. But, $E$ is X-form that could have subjective factors. One circuit could be several different X-forms. The way to form a X-form from a circuit is not unique. We see some examples below.

\begin{exmp}[\bf X-form and Circuit] Some examples of X-form and circuits.\\
{\bf 1: }Suppose $N = 3$. We have a boolean circuit: $C: C(b_1, b_2, b_3) = b_1 \oplus b_2$. This boolean circuit is X-form actually in this way: Let $p_1 = (1, 0, 0), p_2 = (0, 1, 0)$, so $p_1, p_2$ both are base patterns, and one algebraic expression $E(p_1, p_2) = (p_1 \lor p_2) \land \neg (p_1 \land p_2)$, so $E$ is one X-form. We can see $\forall v \in \B^3, E(v) = C(v)$. 

{\bf 2: }Suppose $N = 4$, we have a group of base patterns: $\{p_1, p_2, p_3\}, p_1 = (1,0,0,0), p_2 = (0,1,0,0), p_3 = (0,0,1,1)$, and algebraic expression $E(p_1, p_2, p_3) = p_3 \lor p_1 \lor p_2 \lor (p_1 \land p_2)$. They will form an X-form $E$. Then, this X-form $E$ is equivalent to a boolean circuit: $C: C(b_1, b_2, b_3, b_4) = (b_3 \land b_4) \lor b_1 \lor b_2 \lor (b_1 \land b_2)$. 
\end{exmp}

From the above lemmas, we know that X-forms are equivalent to boolean circuit. Thus, looking for better X-form is equivalent to looking for better boolean circuit. 

FE and PSS provide us a new set of tools for finding better circuit, thus, better X-form.

\subsection*{Learning Strategies by Using Fitting Extremum and PSS}
In \cite{paper2}, we discussed learning dynamics and suggested several learning strategies. As a consequence of such discussions, we showed that deep learning can be explained by the learning strategy called "Embed X-forms into Parameter Space". From its root, this learning strategy needs a lot of human interventions, which is not desirable. In order achieve learning without human interventions, we invented other strategies called: "Squeeze X-form from Inside to Higher Abstraction", and "Squeeze X-form from Inside and Outside to Higher Abstraction". We showed that if we have data that are sufficient to bound and sufficient to support the X-form, the above 2 strategies could realize universal learning (i.e. be able to learning any possible to learn without human interventions).

However, these learning strategies are not good enough, which need huge data (sufficient to bound and sufficient to support often equivalent to huge data) and depend on some capabilities that are still on development. In fact, we know very clearly that these learning strategies are just our first attempt in the study of universal learning machine. They helped us to gain theoretical understanding, but they are not practical. We need better methods. Now, with newly invented tools, i.e. FE and PSS, we can design much better learning strategies.    

Suppose the learning machine is $\mm$, the conceiving space of $\mm$ is $\mathcal{C}$ , the current X-form in $\mathcal{C}$ is $E$.  We also denote the input data as $D = \{ (b_j, o_j) \ |  j = 1, 2, \ldots \}$. In this framework, the learning is: under the driven of input data, the current X-form $E$ is moving to the X-form that we desire. The learning strategy is how to move/change $E$, effectively and efficiently.

Here, we design 2 strategies. Both are based on FE and PSS. The first strategy does learning pure objectively, while the second utilizes subjective view of machine. We discuss 2 strategies separately below. 

Suppose data input are: $D = \{ (b_j, o_j) \ |  j = 1, 2, \ldots \}$, where $b_j \in \B^N$ are base patterns as input. $o_j$ are the value of output should take, but $o_j$ could be empty. If $o_j$ is not empty, $o_j \in \B$. This means we know the output of information processing. If $o_j$ is empty, it means that we do not know (or do not need to know) the output of information processing. If in learning, each $o_j$ is not empty, it is supervised learning. \\

{\bf Learning Strategy -- Objectively Using Fitting Extremum}\\
We can call this strategy as Strategy OF. For Strategy OF, we need to put one requirement on its data input: in data input $D = \{ (b_j, o_j) \ |  j = 1, 2, \ldots, K \}$, $b_j \in \B^N, o_j \in \B$, $o_j$ are not empty, for all $j$. 

We summarize Strategy OF as: 
\begin{enumerate} [topsep=0pt,itemsep=-1ex,partopsep=1ex,parsep=1ex]
\item In this strategy, X-form is actually a boolean circuit. At each step, the current X-form is $E$.
\item At first, the initial X-form is $E_0$, which could be any X-form. Set $E = E_0$, 
\item Start from the first data input: $(b_1, o_1)$.
\item At $J$-th step, $J<K$, data input is $(b_J, o_J)$. Then, first check if $E(b_J) = o_J$. If it is true, this step is done, no need to do further, and go to next step. 
\item If $E(b_J) \neq o_J$, then need to update $E$. The way to update is: To form the sampling set with value $Sv_J = \{ [b_j, o_j] \ |  j = 1, 2, \ldots, J \}$, then do FE on $Sv_J$ to generate circuit $C$, then use this $C$ to replace $E$. 
\item Decide if continue learning. If so, go to next step.
\end{enumerate}

Strategy OF are purely driven by data, i.e. learning machine $\mm$ will do learning objectively according to incoming data. This is why we call it as "objectively using FE". We have following theorem about Strategy OF.

\begin{theorem}[\bf Strategy OF]
Suppose a learning machine $\mm$, and suppose data $D = \{ (b_j, o_j) \ | j = 1, 2, \ldots, K \}$ is used to drive learning, and we are using Strategy OF, if the desired X-form is $E_d$, and the sampling set $Sv_J = \{ [b_j, o_j] \ | j = 1, 2, \ldots, J \}$ is a PSS for $E_d$ for some $J<K$, then, starting from any X-form $E_0$, eventually, $\mm$ will learn $E_d$, i.e. the current X-form $E$ will become the desired X-form $E_d$.        
\end{theorem}
{\bf Proof:}  It is easy to see the proof. Since for some $J<K$, the sampling set $Sv_J$ is a PSS for $E_d$, when we do FE on $Sv_J$, the circuit generated will be $E_d$. That is to say, once the data feed is long enough (i.e. greater than $J$), the current X-form becomes $E_d$. 
\hfill$\blacksquare$

\begin{corollary}
A learning machine $\mm$ with Strategy OF is an universal learning machine.
\end{corollary}
{\bf Proof:} For any given starting X-form $E_0$, and any desired X-form $E_d$, if we give data input that form PSS for $E_d$, then without any human intervention, $\mm$ will learning $E_d$. That is to say, $\mm$ is an universal learning machine. 
\hfill$\blacksquare$

Comparing with other learning strategies we discussed before, the advantage of Strategy OF is very clear: it needs much less data. It only need a data set that including a PSS for the desired X-form, which is much smaller than sufficient to bound and sufficient to support data. This will make learning much better and faster. 

Another advantage is that Strategy OF gives a definitive method to do evolution of current X-form. In other methods we discussed in \cite{paper2}, we only assume some learning capabilities that are still waiting to be realized. With Strategy OF, we are ready to put universal learning machine into practical stage. 

One thing we need to state again: Strategy OF requires the data $o_j$ are not empty. This is a very big restriction. \\

We then turn to another learning strategy. In this strategy, we utilize subjective view of machine, which makes learning better. Compare to pure objective way, subjective way is better in many aspects. One such aspects is: we data $o_j$ could be empty for some $j$. \\

{\bf Learning Strategy -- Subjectively Using Fitting Extremum}\\
We can call this strategy as Strategy SF. In Strategy OF, we use FE and PSS pure objectively, and we require $o_j$ are not empty for all $j$. But, in Strategy SF, we will utilize subjective view of machine in learning, and some $o_j$ could be empty.   

We summarize this strategy as: 
\begin{enumerate} [topsep=0pt,itemsep=-1ex,partopsep=1ex,parsep=1ex]
\item At each step, the current X-form is $E$.
\item At first, the initial X-form is $E_0$, which could be any X-form. Set $E = E_0$.
\item In conceiving space $\mathcal{C}$, maintaining a set of X-forms that are available to be used. Denote this set of X-forms as $\mathcal{X}$. This set $\mathcal{X}$ is super important. When we are looking for X-form to be used, we are looking for X-form only in $\mathcal{X}$. $\mm$ will subjectively maintain this set $\mathcal{X}$ (of course, under the driven of data).  
\item Start from the first data input: $(b_1, o_1)$.
\item At $J$-th step, $J<K$, data input is $(b_J, o_J)$. If $o_j$ is not empty, check if $E(b_J) = o_J$. Then, there are 3 situations: 1) $o_j$ is empty, 2) $o_j$ is not empty, and $E(b_J) = o_J$, 3) $o_j$ is not empty, and $E(b_J) \neq o_J$.  
\item For situation 1), do subjective actions to maintain the set $\mathcal{X}$. 
\item For situation 2), do subjective actions to maintain the set $\mathcal{X}$. Also, keep $o_j$ and information about $E(b_J) = o_J$. 
\item For situation 3),  need to update $E$ to fit the data, first form a sampling set with value as: $Sv_J = \{ [b_i, o_i] \ |  i = 1, 2, \ldots, I_J \}$, where $b_i, o_i$ are pairs of data input: $o_j$ are not empty. $I_J$ is the index of, Then, do FE on $Sv_J$. But, available X-forms are chosen from $\mathcal{X}$. Suppose the circuit $C$ generated by FE on $Sv_J$ over $\mathcal{X}$ is $C$, and the associated X-form from $C$ is $E'$, then use $E'$ to replace the current X-form.
\item Decide if more learning. If so, go to next step.
\end{enumerate}

Strategy SF ultimately is driven by inputing data, but, there are significant subjective actions. This is why we call it as "subjectively using FE". We have following theorem about this strategy.

\begin{theorem}[\bf Strategy SF]
Suppose a learning machine $\mm$, and suppose data $D = \{ (b_j, o_j) \ | j = 1, 2, \ldots, K \}$ is used to drive learning, and we are using Strategy SF to learn, if the desired X-form is $E_d$, and if there is a sampling set $Sv_I = \{ [b_i, o_i] \ | i = 1, 2, \ldots, I \}$ embedded in $D$, and $Sv_I$ is a PSS for $E_d$, then, starting from any X-form $E_0$, eventually, $\mm$ will learn $E_d$, i.e. the current X-form $E$ will become the desired X-form $E_d$.      
\end{theorem}
{\bf Proof:} Suppose the subjective actions in learning is in right direction, so that eventually, $\mathcal{X}$ will have $E_d$ inside it, and sampling set $Sv_I$ is eventually be used. Since there is a sampling set $Sv_I$ embedded in data $D$, and it is a PSS for $E_d$, when we do FE on $Sv_I$, the circuit generated will be $E_d$. That is to say, eventually, the current X-form is $E_d$. 
\hfill$\blacksquare$

\begin{corollary}
A learning machine $\mm$ with Strategy SF is an universal learning machine.
\end{corollary}

By using subjective actions, we are possible to speedup the learning very substantially if these subjective actions are in the right direction (the performance could become worse if the subjective action is not good). So, Strategy SF could learn much faster than Strategy OF. What are subjective actions and how to do subjective actions efficiently actually is big question. We will discuss this in other places.

\section{Discussions}
We make some comments about FE, PSS and learning dynamics.

\begin{enumerate} 
\item FE+PSS (fitting extremum and proper sampling set) are important tools. They are highly related to machine epistemology, i.e. how a machine learns a rule in its environments and how machine represents the learned rule inside itself. FE+PSS tells us: the rule is in fact inside a set of data (data contains PSS), and if machine keeps looking better representations (X-form) with least cost (fewest nodes), eventually, machine learns the rule fully. This has very strong epistemological meaning. It is worth to do deep study. We will discuss this issue in other place.

\item We proved the fundamental relationship between PSS and complexity of boolean circuit. This gives us a strong tool to study computational complexity. We will explore this in the next study. This fundamental relationship between PSS and computational complexity actually reflect the intrinsic relationship between learning and computational complexity, and such intrinsic relationship is the very core of learning.

\item FE reveals why generalization can be achieved in mechanical learning. From view of FE, we will see generalization very naturally, no longer with surprise. 

\item With FE+PSS, and learning strategies OF and SF, universal learning machine is no longer just theoretically true, but is in practical stage. Our previous papers discussed other learning strategies. But, the Strategy OF and Strategy SF are much different, and much better. Strategy OF and SF are ready to be used in engineering practice.

\item In order to use Strategy OF and Strategy SF, we need to do FE efficiently. It can be done. Our next research project will be how to do FE efficiently and effectively. 

\item In Strategy SF, we can do subjective actions, which can help learning to speedup. This will be one very fruitful research area.

\end{enumerate}


\newpage
\section*{Appendix}
In appendix, we want to prove the 2 lemmas and 2 theorems stated in section 4, i.e. Expansion of Functions, PSS implies Circuit and Circuit implies PSS. 

We first put PSS implies circuit below. \\

{\bf PSS implies Circuit: } {\it
If $f$ is a boolean function $f: \B^N \to \B$, and $S \subset \B^N$ is a PSS for $f$, and $|S|$ is the size of PSS, then there is a circuit $C$ expresses $f$ and $d(C) < N|S|$. } 

{\bf Proof: } Now, let $K = |S|$, and $S = \{v_1, v_2, \ldots, v_K\}$. Let $C_{v_j}$ be circuit to express $v_j$, where $,j=1, 2, \ldots, K$. So, if $x = v_j$, $C_{v_j}(x) = 1$ and if $x \neq v_j$, $C_{v_j}(x) = 0$. Using them, we form one circuit $C_f = (\ldots(s_1 C_{v_1} \lor s_2 C_{v_2}) \lor \ldots \lor s_K C_{v_K})$, where $s_j$ are: if $f(v_j) = 1, s_i = id, otherwise, s_i = \neg$. It is clear, $\forall x \in S, C_f(x) = f(x)$, i.e. circuit $C_f$ is fitting with $S$. Also, we can see $C_f$ has $K-1$ "$\lor$" nodes, and each $C_{v_j}$ has $N-1$ "$\land$" nodes, so  $d(C_f) = K-1 + K(N-1) = KN-1 < N|S|$. This is to say, there is a circuit $C_f$ fitting with $S$ and $d(C_f) < N|S|$. 

Therefore, if a circuit $C$ is the circuit generated by fitting extremum from $f$ and $S$, since $S$ is PSS.  $C$ should expresses $f$. And, $d(C) \le d(C_f) < N|S|$.    
\hfill$\blacksquare$

For circuit implies PSS, we need some lemmas first.

Suppose $C$ is a circuit and $w$ is one node of $C$, then for each vector $b \in \B^N$, $w$ will take some value accordingly. We will call this value as the value at node $w$ for input $b$, denote as $w(b)$. If the $w$ is the top node, then $w(b)$ is the value of the circuit for input $b$, i.e. $C(b)=w(b)$. Here is a lemma that tells us about the values at nodes of circuit.\\

\begin{lemma}[\bf Value at Node]
Suppose $f: \B^N \to \B$ is a boolean function, $C$ is a boolean circuit expressing $f$, and $d(C)$ reaches minimum, then, for any node $w$ in $C$, the values at the 2 nodes $w_L, w_R$ directly underneath $w$ must satisfies the following rules: for each type of connection configurations (totally 8 types), there must have inputs $b_1, b_2, b_3 \in \B^N$ so that $(w_L(b_1), w_R(b_1))$, $(w_L(b_2), w_R(b_2))$, $(w_L(vb_3), w_R(b_3))$ takes values specified below.
\end{lemma}
{\bf Proof: } There are 8 connection configurations as below: 
$\begin{bmatrix}\lor & \lor\end{bmatrix}$, 
$\begin{bmatrix}\lor & \lor\neg\end{bmatrix}$,
$\begin{bmatrix}\lor\neg & \lor\end{bmatrix}$,
$\begin{bmatrix}\lor\neg & \lor\neg\end{bmatrix}$
$\begin{bmatrix}\land & \land\end{bmatrix}$,
$\begin{bmatrix}\land & \land\neg\end{bmatrix}$,
$\begin{bmatrix}\land\neg & \land\end{bmatrix}$,
$\begin{bmatrix}\land\neg & \land\neg\end{bmatrix}$.

First consider $\begin{bmatrix}\lor & \lor\end{bmatrix}$. We want to show: there must have 3 inputs $b_1, b_2, b_3 \in \B^N$ so that $(w_L(b_1), w_R(b_1)) = (0,0), (w_L(b_2), w_R(b_2))=(1,0), (w_L(b_3), w_R(b_3))=(0,1)$.

First, if there no $b \in \B^N$ so that $(w_L(b), w_R(b)) = (0,0)$, then, due to the connection configuration of node $w$, the value at $w$ is always 1. In this case, the circuit $C$ can be simplified to another circuit $C'$ and $\forall b\in \B^N, C(b)=C'(b)$, and $d(C') < d(C)$. This is a contradiction to $d(C)$ reaches minimum. So, there is at least one $b_1 \in \B^N$ so that  $(w_L(b), w_R(b)) = (0,0)$.

Next, suppose there is no $b \in \B^N$ so that $(w_L(b_2), w_R(b_2))=(1,0)$, it means: for any $b \in \B^N$, there are only 3 possibilities: $(w_L(b), w_R(b))=(0,0)$ or $(1,1)$ or $(0,1)$. So, we see the value at $w(b)$ equals value at $w_R$, i.e. $w(b) = w_R(b), \forall b \in \B^N$. So, we can eliminate node $w_L$ without modifying value of $w$. In this case, the circuit $C$ can be simplified to another circuit $C'$ and $\forall b\in \B^N, C(b)=C'(b)$, and $d(C') < d(C)$. This is a contradiction to $d(C)$ reaches minimum. So, there is at least one $b_2 \in \B^N$ so that  $(w_L(b_2), w_R(b_2)) = (1,0)$.

Next, by exactly same argument, we know, there is at least one $b_3 \in \B^N$ so that  $(w_L(b_3), w_R(b_3)) = (0,1)$.  

Then, consider $\begin{bmatrix}\land & \land\end{bmatrix}$. By the very similar arguments as above, we can show: there at least one $b_1 \in \B^N$ so that $(w_L(b_1), w_R(b_1)) = (1,1)$, at least one $b_2 \in \B^N$ so that $(w_L(b_2), w_R(b_2)) = (1,0)$, at least one $b_3 \in \B^N$ so that $(w_L(b_3), w_R(b_3)) = (0,1)$. We skip the details.

For all other types, we can have the similar arguments and get similar results. We skip the details. We list all results for all connection configurations below. 

For $\begin{bmatrix}\lor & \lor\end{bmatrix}$, there are 3 inputs so that the value at nodes $w_L$ and $w_R$ are: (0,0), (1,0), (0,1). \\
For $\begin{bmatrix}\lor & \lor\neg\end{bmatrix}$, there are 3 inputs so that the value at nodes $w_L$ and $w_R$ are: (0,1), (1,1), (0,0). \\
For $\begin{bmatrix}\lor\neg & \lor\end{bmatrix}$, there are 3 inputs so that the value at nodes $w_L$ and $w_R$ are: (1,0), (0,0), (1,1). \\
For $\begin{bmatrix}\lor\neg & \lor\neg\end{bmatrix}$, there are 3 inputs so that the value at nodes $w_L$ and $w_R$ are: (1,1), (0,1), (1,0). \\
For $\begin{bmatrix}\land & \land\end{bmatrix}$, there are 3 inputs so that the value at nodes $w_L$ and $w_R$ are: (1,1), (1,0), (0,1). \\
For $\begin{bmatrix}\land & \land\neg\end{bmatrix}$, there are 3 inputs so that the value at nodes $w_L$ and $w_R$ are: (1,0), (1,1), (0,0).\\
For $\begin{bmatrix}\land\neg & \land\end{bmatrix}$, there are 3 inputs so that the value at nodes $w_L$ and $w_R$ are: (0,1), (0,0), (1,1). \\
For $\begin{bmatrix}\land\neg & \land\neg\end{bmatrix}$, there are 3 inputs so that the value at nodes $w_L$ and $w_R$ are: (0,0), (0,1), (1,0).
\hfill$\blacksquare$

By observing the results of Lemma 6.1, we can see something very interesting and useful. First consider at $w$ the connection configuration is $\begin{bmatrix}\lor & \lor\end{bmatrix}$, then, we have $b_1, b_2, b_3$ so that the values at $w_L, w_R$ are: $(0,0),(1,0),(0,1)$. So, if we let $S_L = \{b_1, b_2\}, S_R =  \{b_1, b_3\}$, then $\forall x \in S_L, w_R(x)=0$, and $\forall x \in S_R, w_L(x)=0$. Also, $\forall x \in S_L, w(x) = w_L(x)$, and  $\forall x \in S_R, w(x) = w_R(x)$. 

Then, consider at $w$ the connection configuration is $\begin{bmatrix}\land & \land\end{bmatrix}$, then, we have $b_1, b_2, b_3$ so that the values at $w_L, w_R$ are: $(1,1),(1,0),(0,1)$. So, if we let $S_L = \{b_1, b_2\}, S_R =  \{b_1, b_3\}$, then $\forall x \in S_L, w_R(x)=1$, and $\forall x \in S_R, w_L(x)=1$. Also, $\forall x \in S_L, w(x) = w_L(x)$, and  $\forall x \in S_R, w(x) = w_R(x)$.

For all other type of connection configuration, we have similar results. These results are important for later usage. 
\\

\begin{lemma}[\bf Expansion of Sampling]
Suppose $f: \B^N \to \B$ is a boolean function, and $S \subset \B^N$ is a PSS for $f$. If we expand the sampling, i.e. let $b \in \B^N, b \notin S$, and $S' = S \cup \{b\}$, and we set the value on $b$ different than $f(b)$. If $D$ is a circuit fits with $S$, and $d(D)$ reaches minimum, and $D'$ is a circuit generated from FE on $S'$, then $d(D) < d(D')$. 
\end{lemma}
{\bf Proof: } Since $S$ is PSS, and $d(D)$ reaches the minimum, so circuit $D$ must expresses $f$. Now, let $D'$ be a circuit generated from FE on $S'$. Since $D'$ fits with $S'$, so fits with $S$, by definition of PSS, $d(D) \le d(D')$. Further, if $d(D') = d(D)$, which means the circuit $D'$ fits with $S$ and the number of nodes reaches minimum. Since $S$ is PSS, it means such circuit $D'$ must expresses $f$. However, the value of $D'$ on $b$ is different than $f(b)$ as $D'$ fits with $S'$. This is a contradiction. The contradiction tells $d(D') = d(D)$ is wrong. Thus, we must have $d(D) < d(D')$.  
\hfill$\blacksquare$

We can have weaker version.\\

\begin{lemma}[\bf Expansion of Sampling, Weaker]
Suppose $S \subset \B^N$ is a sampling set, not necessarily a PSS. And $Sv=\{[s, v] \ | \ s \in S, v = 0 \text{ Or } 1 \}$ is a sampling set with value over $S$. $D$ is a circuit fits with $Sv$, and $d(D)$ reaches minimum.  Suppose $b \in \B^N, b \notin S$, we expand sampling set with value as $Sv' = Sv \cup \{[b, v]\}$, where $v$ is such a value: $v$ is different than $D(b)$. Suppose $D'$ is a circuit generated from FE on $Sv'$, then $d(D) < d(D')$. 
\end{lemma}
{\bf Proof: } We want to use the the above lemma (i.e. Expansion of Sampling). The problem is: $S$ is not necessarily a PSS. So, we need to make some additional arguments. Define a boolean function: $f(t) = D(t), \forall t \in \B^N$. If $S$ is a PSS for $f$, then we can apply above lemma and the proof is done. If $S$ is not PSS for $f$, we can add some points to $S$ to get a sampling set $S^*$, so that $S^*$ becomes a PSS of $f$. This surely can be done. In this case, $D$ is still a circuit generated from FE on $S^*$. Then, we can apply above lemma, and proof is done.
\hfill$\blacksquare$

The above 2 lemmas tells us this: if the sampling expands, then the circuit generated by FE on the sampling will expands as well. That is to say, for a more complicated sampling, the circuit generated from FE on it must be bigger, with more nodes. This is one fundamental fact that plays important role. 

Next, we want show how to join PSSs to form new PSS. \\

\begin{lemma}[\bf Join PSS]
If $f: \B^N \to \B$ is a boolean function, and circuit $C$ expresses $f$, also $d(C)$ reaches minimum. $C$ must be in such a form: $C=L \circ R$, where $\circ$ is the connection configuration of top node (there are 8 types, see Lemma 6.1), and $L, R$ are 2 sub-circuits of $C$. Suppose $S_L, S_R\subset  \B^N$ are 2 sets with property: 1) $S_L$ is a PSS of $L$ and $S_R$ is a PSS of $R$, 2) $\forall x \in S_L, f(x)=L(x)$ and $\forall x \in S_R, f(x)=R(x)$, then the set $S = S_L \cup S_R$ is a PSS of $f$.
\end{lemma}
{\bf Proof:} We can think a process to seek circuit $D$ that fits $S_L \cup S_R$ while keep $d(D)$ to be lowest. We can start from $S_L$, and do FE on $S_L$. Suppose we get circuit $D_L$. Since $S_L$ is PSS for $L$, must $\forall x \in \B^N, D_L(x) = L(x)$. The next step is to consider modify circuit $D_L$ to get circuit $D_{L+R}$ so that $D_{L+R}$ will keep $D_L(x), \forall x\in S_L$, and $D_{L+R}$  fits with $S_R$, and also make $d(D_{L+R})$ to be lowest. The only possible choice is: $D_{L+R} = D_L \circ D_R$, whiere $D_R$ is a circuit from FE on $S_R$. Since $S_R$ is PSS for $R$, must $\forall x \in \B^N, D_R(x) = R(x)$. Thus, $\forall x \in \B^N, D_{L+R}(x) = D_L(x) \circ D_R(x) = C(x)=f(x)$. This tells us that $S_L \cup S_R$ is PSS of $f$. 
\hfill$\blacksquare$

This lemma tells us one very essential property PSS: it must grasp the characteristics of each branch, and can distingish branch from each other. Using this property, we know how to pick up PSS from a circuit. \\

{\bf Pick up PSS by using circuit: } We are going to pick up a sampling set from a given circuit. Suppose $f: \B^N \to \B$ is a boolean function, circuit $C$ expresses $f$, and $d(C)$ reaches minimum. 

We are going to pick up sampling by using Lemma 6.4, which tells us how to join PSSs of 2 branches together to form a PSS. 

We first consider the simplest circuit, the circuits with height 1. In order to make writing easier, we consider $\B^4$. Such circuit $C$ must in form: $C= L\circ R$, where $\circ$ is one of 8 types of connection configuration shown in Lemma 6.1, and $L, R$ are 2 sub-circuits of $C$. In this case, due to height 1, must $L=b_i, R=b_j, i,j=1,\ldots,4, i \ne j$. We can see some vectors in $\B^4$ below.
$$
b_1 = \begin{bmatrix} 
       1 \\ 0 \\ 0 \\ 0 
\end{bmatrix},
b_2 = \begin{bmatrix} 
       0 \\ 1 \\ 0 \\ 0 
\end{bmatrix},
b_3 = \begin{bmatrix} 
       0 \\ 0 \\ 0 \\ 0 
\end{bmatrix},
b_4 = \begin{bmatrix} 
       1 \\ 1 \\ 0 \\ 0 
\end{bmatrix},
b_5 = \begin{bmatrix} 
       1 \\ 0 \\ 0 \\ 1 
\end{bmatrix},
b_6 = \begin{bmatrix} 
       0 \\ 1 \\ 0 \\ 1 
\end{bmatrix},
b_7 = \begin{bmatrix} 
       0 \\ 0 \\ 0 \\ 1 
\end{bmatrix},
b_8 = \begin{bmatrix} 
       1 \\ 1 \\ 0 \\ 1 
\end{bmatrix},
b_9 = \begin{bmatrix} 
       1 \\ 0 \\ 1 \\ 1 
\end{bmatrix}
$$
As Lemma 6.4 tells us, we can find PSS for $L$, and PSS for $R$, and the satisfies certain condition, then, join these 2 PSSs, we get PSS for $C$. Consider one example: $C= L\land R, L = b_1, R = b_2$. It is easy to see $S_L = \{b_1, b_3 \}$ is a PSS of $L$, and $S_L = \{b_5, b_7 \}$ is a PSS of $L$, and $S_L = \{b_2, b_4 \}$ is a PSS of $L$. Also, there are several choice for PSS of $R$. However, the sets $S_L = \{b_2, b_4 \}$ and $S_R = \{b_1, b_4 \}$ have properties: $\forall x \in S_L, C(x)=L(x)$, and $\forall x \in S_R, C(x)=R(x)$. This property is essential. With it, by Lemma 6.4, $S_L \cup S_R$ is PSS for $C$. 

This is for top node as $\land$. But, we can do exactly same for other type of node. See Lemma 6.1. This is how to pick up PSS from a circuit with height 1. Moreover, $S_L = \{b_6, b_8 \}$ and $S_R = \{b_5, b_8 \}$ can be used to form PSS for $C$.

For height as 1, clearly, $|S|=3$, and $d(C)=1$. So, $|S| \le 3d(C)$.

For a circuit $C$ expressing $f$ and $d(C)$ reaches minimum, any sub-circuits $D$ of $C$ expresses a boolean function, we use $\forall x\in \B^4 D(x)$ to represent this sub-circuit. Easy to see, $d(D)$ reaches minimum. So, we pick up sampling set in this way: For 2 branches of $D$, $L, R$, we can have $S_L$ and $S_R$,  with this property: $S_L$ is PSS for $L$, and $S_R$ is PSS for $R$, and $\forall x \in S_L, C(x)=L(x)$, and $\forall x \in S_R, C(x)=R(x)$. Then, $S=S_L \cup S_R$ will be PSS for $D$, and $|S| \le 3d(D)$. 

We do this for all sub-circuits of $C$, then finally reache to the top of $C$. In this way, we eventally get 2 sampling sets $S_L$ and $S_R$ for $C$, so that $S=S_L \cup S_R$ is PSS for $C$, and $|S| \le 3d(C)$. 
\hfill$\blacksquare$
\\

The above process already shows: circuit implies PSS. We just state this again below.

{\bf Circuit implies PSS: } {\it 
$f: \B^N \to \B$ is a boolean function, a circuit $C$ expresses $f$, and $d(C)$ reaches minimum, then, we can pick up a sampling set $S$ so that $S$ is PSS of $f$, and $|S| \le 3d(C)$. } 

We consider some simple example of how to pick PSS from a circuit.\\

\begin{exmp}[\bf Example of PSS]
1. Consider $\B^3$, and one very simple boolean function: $f: \B^3 \to \B, f(b_1, b_2, b_3)=b_1\lor b_2$. Circuit $C=b_1 \lor b_2$ expresses $f$, and $d(C)$ reaches minimum 1. Thus, $C=L\lor R, L=b_1, R=b_2$. Very clear, following sets:
$$
S_L = \left\{ 
\begin{bmatrix} 
       0 \\       0 \\ 0
\end{bmatrix},
\begin{bmatrix} 
      1 \\      0 \\ 0
\end{bmatrix}
\right\}, \quad
S_R = \left\{
\begin{bmatrix} 
      0 \\      0 \\ 0
\end{bmatrix},
\begin{bmatrix} 
      0 \\      1 \\ 0
\end{bmatrix}
\right\}
$$
are PSS for $L$ and PSS for $R$. Also, very clear, if $x \in S_L$, $R(x)=0$, and if $x \in S_R$, $L(x)=0$. By the Lemma 6.5, the set $S=S_L \cup S_R$ is PSS for $C$. Note, $|S|=3$ (since there is some overlapping of $S_L$ and $S_R$), so $|S| \le 3d(C)=3$. 

2. Still consider $\B^3$, and one boolean function: $f: \B^3 \to \B, f(b_1, b_2, b_3)=(b_1\lor b_2) \land b_3$. So, circuit $C=(b_1\lor b_2) \land b_3 = L \land R$ expresses $f$, and $d(C)$ reaches minimum 2. Note, $C$ has height 2. Now $C=L\land R$, where $L=b_1\lor b_2, R=b_3$ are 2 sub-circuit of $C$. Note, here, the sub-circuit $L$ is the circuit in example 1 above, and we already know one PSS of $L$. We have following sets: 
$$
S_L = \left\{
\begin{bmatrix} 
       0 \\       0 \\       1
\end{bmatrix},
\begin{bmatrix} 
      1 \\      0 \\      1
\end{bmatrix},
\begin{bmatrix} 
      0 \\      1 \\      1
\end{bmatrix} \right\}, \quad 
S_R = \left\{
\begin{bmatrix} 
       1 \\       0 \\       0
\end{bmatrix},
\begin{bmatrix} 
      1 \\      0 \\      1
\end{bmatrix} \right\}
\quad \ Or \  
S_R = \left\{
\begin{bmatrix} 
       1 \\       1 \\       0
\end{bmatrix},
\begin{bmatrix} 
      1 \\      1 \\      1
\end{bmatrix} \right\}
$$
We know $S_L$ is PSS of $L$, and $S_R$ is PSS of $R$. And, they have the property: if $x \in S_L$,  $R(x)=1$, and if $x \in S_R$, $L(x)=1$. By the Lemma 6.3, the set $S=S_L \cup S_R$ is PSS for $C$. Note, $|S|=4$ (since there is some overlapping of $S_L$ and $S_R$), or $|S|=5$, so $|S| \le 3d(C)=6$.  It is worth to note this: here $S_L$ are PSS of $L=b_1\lor b_2$, compare to $f$ and $S$ in example 1. 

3. Consider circuit with height 3. We have $f: \B^4 \to \B, f(b_1, b_2, b_3, b_4)=(b_1\land \neg b_2) \lor ((\neg b_1 \land b_2) \land (b_3 \lor b_4))$. Circuit $C=(b_1\land \neg b_2) \lor ((\neg b_1 \land b_2) \land (b_3 \lor b_4))$ expresses $f$. So $C$ is in the form: $C = L \lor R$, where $L= b_1\land \neg b_2$, $R= (\neg b_1 \land b_2) \land (b_3 \lor b_4)$. We have $S_L, S_R$:
$$
S_L = \left\{
\begin{bmatrix} 
       1 \\ 0 \\ 0 \\ 0 
\end{bmatrix},
\begin{bmatrix} 
       1 \\ 1 \\ 0 \\ 0 
\end{bmatrix},
\begin{bmatrix} 
       0 \\ 0 \\ 0 \\ 0 
\end{bmatrix} \right\}, \quad 
S_R = \left\{
\begin{bmatrix} 
0 \\ 1 \\ 0 \\ 0
\end{bmatrix},
\begin{bmatrix} 
0 \\ 1 \\ 1 \\ 0
\end{bmatrix},
\begin{bmatrix} 
0 \\ 1 \\0 \\ 1
\end{bmatrix},  
\begin{bmatrix} 
0 \\1 \\ 1 \\ 1
\end{bmatrix},
\begin{bmatrix} 
0 \\0\\1\\1
\end{bmatrix},
\begin{bmatrix} 
1\\1\\1\\1
\end{bmatrix} \right\}
$$
Easy to see, if $x \in S_L$, $R(x)=0$, and if $x \in S_R$, $L(x)=0$. And, $S_L$ is PSS of $L$, and $S_R$ is PSS of $R$. By the Lemma 6.3, the set $S=S_L \cup S_R$ is PSS for $C$. Note, $|S|=9 < 3d(C)=15$.
\end{exmp}

\end{document}